\documentclass[11pt,a4paper]{article}
\usepackage[hyperref]{acl2020}
\usepackage{times}
\usepackage{latexsym}

\usepackage{microtype}

\aclfinalcopy 
\def\aclpaperid{3104} 

\usepackage{amsmath}
\usepackage{cleveref}
\usepackage{amsfonts}

\newcommand{\keyword}[1]{\textit{#1}}

\usepackage{subcaption}

\usepackage{tabu}
\usepackage{booktabs}
\usepackage{multirow}

\usepackage{mathtools}
\DeclarePairedDelimiter{\abs}{\lvert}{\rvert}

\usepackage{bm}

\def\eqref#1{equation~\ref{#1}}

\def\1{\bm{1}}

\def\rvw{{\mathbf{w}}}

\def\rmW{{\mathbf{W}}}

\def\vw{{\bm{w}}}

\def\mW{{\bm{W}}}
\def\mX{{\bm{X}}}

\DeclareMathAlphabet{\mathsfit}{\encodingdefault}{\sfdefault}{m}{sl}
\SetMathAlphabet{\mathsfit}{bold}{\encodingdefault}{\sfdefault}{bx}{n}

\def\gG{{\mathcal{G}}}

\def\sS{{\mathbb{S}}}

\def\sW{{\mathbb{W}}}

\DeclareMathOperator*{\argmax}{arg\,max}

\usepackage[warn]{textcomp}

\usepackage{color,soul}

\newcommand{\glTOen}{gl\textrightarrow en}
\newcommand{\skTOen}{sk\textrightarrow en}
\newcommand{\arTOen}{ar\textrightarrow en}
\newcommand{\enTOvi}{en\textrightarrow vi}
\newcommand{\enTOde}{en\textrightarrow de}
\newcommand{\enTOar}{en\textrightarrow ar}
\newcommand{\IWSLT}{IWSLT\,\textquotesingle 15}

\usepackage{tikz}
\def\checkmark{\tikz\fill[scale=0.4](0,.35) -- (.25,0) -- (1,.7) -- (.25,.15) -- cycle;}

\title{Masked Language Model Scoring}

\author{Julian Salazar$^{\spadesuit}$ \quad Davis Liang$^\spadesuit$ \quad Toan Q. Nguyen$^\diamondsuit$\thanks{\ \ Work done during an internship at Amazon AWS AI.} \quad Katrin Kirchhoff$^\spadesuit$\\
  $^\spadesuit$\,Amazon AWS AI, USA\\
  $^\diamondsuit$\,University of Notre Dame, USA\\
  {\tt \{julsal,liadavis,katrinki\}@amazon.com, tnguye28@nd.edu} \\
}

\date{}

\begin{document}
\maketitle
\begin{abstract}
Pretrained masked language models (MLMs) require finetuning for most NLP tasks. Instead, we evaluate MLMs out of the box via their \keyword{pseudo-log-likelihood scores} (PLLs), which are computed by masking tokens one by one. We show that PLLs outperform scores from autoregressive language models like GPT-2 in a variety of tasks. By rescoring ASR and NMT hypotheses, RoBERTa reduces an end-to-end LibriSpeech model's WER by 30\% relative and adds up to +1.7 BLEU on state-of-the-art baselines for low-resource translation pairs, with further gains from domain adaptation. We attribute this success to PLL's unsupervised expression of linguistic acceptability without a left-to-right bias, greatly improving on scores from GPT-2 (+10 points on island effects, NPI licensing in BLiMP). One can finetune MLMs to give scores without masking, enabling computation in a single inference pass. In all, PLLs and their associated \textit{pseudo-perplexities} (PPPLs) enable plug-and-play use of the growing number of pretrained MLMs; e.g., we use a single cross-lingual model to rescore translations in multiple languages. We release our library for language model scoring at \url{https://github.com/awslabs/mlm-scoring}.
\end{abstract}

\section{Introduction}

BERT \cite{Devlin2019} and its improvements to natural language understanding have spurred a rapid succession of contextual language representations (\citealp{Yang2019, Liu2019}; \textit{inter alia}) which use larger datasets and more involved training schemes. Their success is attributed to their use of bidirectional context, often via their \keyword{masked language model} (MLM) objectives. Here, a token $\vw_t$ is replaced with \texttt{[MASK]} and predicted using all past and future tokens $\mW_{\backslash t} := (\vw_1, \dotsc, \vw_{t-1}, \vw_{t+1}, \dotsc, \vw_{\abs{\mW}})$.

\begin{figure}[ht!]
\centering
\includegraphics[width=0.75\columnwidth,trim={0 1.4cm 0 0cm},clip]{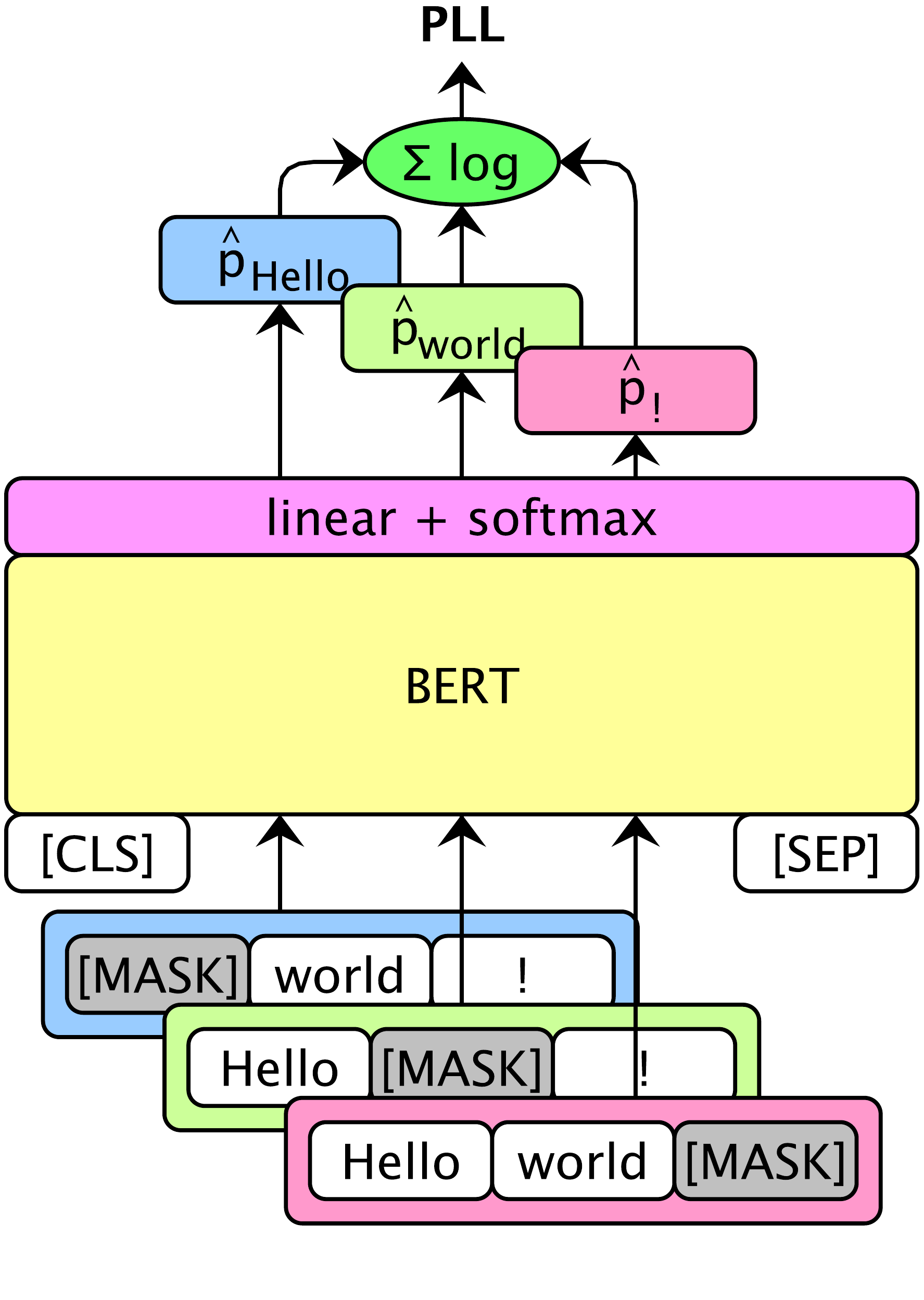}
\caption{To score a sentence, one creates copies with each token masked out. The log probability for each missing token is summed over copies to give the pseudo-log-likelihood score (PLL). One can adapt to the target domain to improve performance, or finetune to score without masks to improve memory usage.}
\label{fig:pll}
\end{figure}

In contrast, conventional language models (LMs) predict $\vw_t$ using only past tokens $\mW_{< t} := (\vw_1, \dotsc, \vw_{t-1})$. However, this allows LMs to estimate log probabilities for a sentence $\mW$ via the chain rule ($\log P_{\text{LM}}(\mW) = \sum_{t=1}^{\abs{\mW}} \log P_{\text{LM}}(\vw_t \mid \mW_{< t})$), which can be used out of the box to rescore hypotheses in end-to-end speech recognition and machine translation \cite{Chan2016, Gulcehre2015}, and to evaluate sentences for linguistic acceptability \cite{Lau2017}.

Our work studies the corresponding \keyword{pseudo-log-likelihood scores} (PLLs) from MLMs \cite{Wang2019}, given by summing the conditional log probabilities $\log P_{\text{MLM}}(\vw_t \mid \mW_{\backslash t})$ of each sentence token \cite{Shin2019}. These are induced in BERT by replacing $\vw_t$ with $\texttt{[MASK]}$ (\Cref{fig:pll}). Let $\Theta$ denote our model's parameters. Our score is
\begin{align*}
    \text{PLL}(\mW) := \sum_{t=1}^{\abs{\mW}} \log P_{\text{MLM}}(\vw_t \mid \mW_{\backslash t}; \Theta).
\end{align*}
PLLs and their corresponding \keyword{pseudo-perplexities} (PPPLs) (\Cref{ssec:pppl-intro}) are intrinsic values one can assign to sentences and corpora, allowing us to use MLMs in applications previously restricted to conventional LM scores. Furthermore, we show that one can finetune BERT to compute PLLs in a single, non-recurrent inference pass (\Cref{ssec:maskless-scoring}).

Existing uses of pretrained MLMs in sequence-to-sequence models for automatic speech recognition (ASR) or neural machine translation (NMT) involve integrating their weights \cite{Clinchant2019} or representations \cite{Zhu2020} into the encoder and/or decoder during training. In contrast, we train a sequence model independently, then rescore its $n$-best outputs with an existing MLM. For acceptability judgments, one finetunes MLMs for classification using a training set \cite{Warstadt2019cola, Devlin2019}; instead, PLLs give unsupervised, relative judgements directly.

In \Cref{sec:seq2seq}, we show that scores from BERT compete with or even outperform GPT-2 \cite{Radford2019}, a conventional language model of similar size but trained on more data. Gains scale with dataset and model size: RoBERTa large \cite{Liu2019} improves an end-to-end ASR model with relative WER reductions of 30\%, 18\% on LibriSpeech \textit{test-clean}, \textit{test-other} respectively (with further gains from domain adaptation), and improves state-of-the-art NMT baselines by up to +1.7 BLEU on low-resource pairs from standard TED Talks corpora. In the multilingual case, we find that the pretrained 15-language XLM \cite{Lample2019} can concurrently improve NMT systems in different target languages.

In \Cref{sec:analysis}, we analyze PLLs and propose them as a basis for other ranking/scoring schemes. Unlike log probabilities, PLL's summands are more uniform across an utterance's length (no left-to-right bias), helping differentiate fluency from likeliness. We use PLLs to perform unsupervised acceptability judgments on the BLiMP minimal pairs set \citep{Warstadt2019}; BERT and RoBERTa models improve the state of the art (GPT-2 probabilities) by up to 3.9\% absolute, with +10\% on island effects and NPI licensing phenomena. Hence, PLLs can be used to assess the linguistic competence of MLMs in a supervision-free manner.

\section{Background}
\label{sec:background}

\subsection{Pseudolikelihood estimation}
\label{ssec:pseudolikelihood}

Bidirectional contextual representations like BERT come at the expense of being ``true'' language models $P_{\text{LM}}(\mW)$, as there may appear no way to generate text (sampling) or produce sentence probabilities (density estimation) from these models. This handicapped their use in generative tasks, where they at best served to bootstrap encoder-decoder models \cite{Clinchant2019,Zhu2020} or unidirectional LMs \cite{WangLiSmola2019}. 

However, BERT's MLM objective can be viewed as stochastic \textit{maximum pseudolikelihood estimation} (MPLE) \cite{Wang2019, Besag1975} on a training set $\sW$, where $\{\rvw_t\}_{t=1}^{\abs{\mW}}$ are random variables in a fully-connected graph. This approximates conventional MLE, with MLM training asymptotically maximizing the objective:
\begin{align*}
\mathcal{J}_{\text{PL}}(\Theta; \sW) = \frac{1}{\abs{\sW}}\sum_{\mW \in \sW} \text{PLL}(\mW; \Theta).
\end{align*}
In this way, MLMs learn an underlying joint distribution whose conditional distributions $\rvw_t \mid \mW_{\backslash t}$ are modeled by masking at position $t$. We include a further discussion in \Cref{ssec:gen-model}.

This enabled text generation with BERT via Gibbs sampling, leading to the proposal (but not evaluation) of a related quantity, the sum of logits, for sentence ranking \citep{Wang2019}. More recent work \citep{Shin2019} extended past research on future-conditional LMs in ASR (\Cref{sec:related-work}) with deeply-bidirectional self-attentive language models (bi-SANLMs). They trained shallow models from scratch with the $\texttt{[MASK]}$ scoring method, but did not relate their work to pseudolikelihood and fluency, which provide a framework to explain their success and observed behaviors.

Experimentally, we extend both works by evaluating pretrained models, domain adaptation, and usage in NMT and multilingual settings (\Cref{sec:results}), along with acceptability judgements and PLL's intrinsic numerical properties (\Cref{sec:analysis}).

\subsection{{\normalfont\texttt{[MASK]}}less scoring}
\label{ssec:maskless-scoring}

A practical point unaddressed in both works is that computing PLLs from an MLM requires a sentence copy for each position, making the number of inference passes dependent on length (though these can be parallelized). The cost of a softmax is also incurred, which is dependent on vocabulary size $V$; together this gives $\mathcal{O}(\abs{\mW} \cdot V)$. We propose reducing this to $\mathcal{O}(1)$ by training a network $q$ with parameters $\Theta_{S}$ to match BERT's PLLs without \texttt{[MASK]} tokens:
\begin{equation*}
	\abs{\text{PLL}(\mW) - q(\mW; \Theta_{S})}^2.
\end{equation*}
We propose finetuning $q$ from the pretrained MLM directly (i.e., initializing $\Theta_{S}$ with $\Theta$), via regression over the \texttt{[CLS]} token (\Cref{fig:maskless}):

\begin{figure}[ht!]
\centering
\includegraphics[width=0.85\linewidth,trim={0 4cm 0 0},clip]{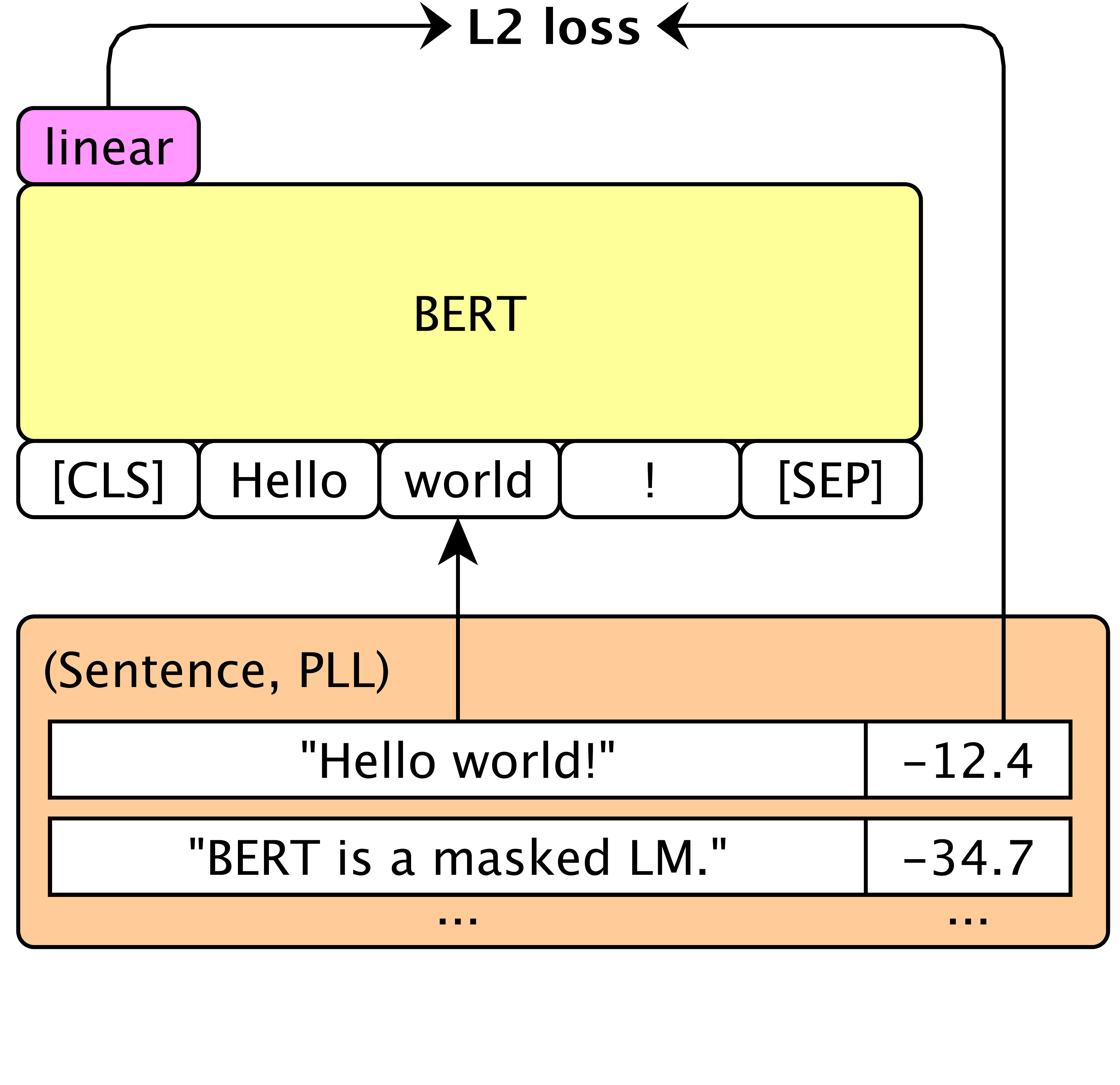}
\caption{We learn a linear map after the \texttt{[CLS]} token, supervised by the PLLs from the pretrained MLM.}
\label{fig:maskless}
\end{figure}

More generally, one could use any student model $q$, as in knowledge distillation \cite{Hinton2014}. Here, the teacher gives individual token probabilities ($\abs{\mW}$ inference passes) while the student approximates their sum (one inference pass). This is reminiscent of distilling an autoregressive teacher to a parallel student, as in the case of WaveNet \cite{Oord2018}.  Other \texttt{[MASK]}less bidirectional models like XLNet \cite{Yang2019} can also give PLLs; we leave this to future work.

\subsection{Pseudo-perplexity}
\label{ssec:pppl-intro}

Analogous to conventional LMs, we propose the \textit{pseudo-perplexity} (PPPL) of an MLM as an intrinsic measure of how well it models a corpus of sentences $\mathbb{W}$. Let $N$ denote the number of tokens in the corpus. Then a model's PPPL on $\mathbb{W}$ is
\begin{align*}
    \text{PPPL}(\mathbb{W}) := \exp\left(- \frac{1}{N} \sum_{\mW \in \mathbb{W}} \text{PLL}(\mW)\right).
\end{align*}
Past work \citep{Chen2017biLM} also computed this quantity with bi-RNNLMs for ASR, although such models are not deeply bidirectional like self-attentive MLMs (see \Cref{sec:related-work}).

These PPPLs can be used in lieu of perplexities. For example, during domain adaptation, one can perform early stopping with respect to development PPPL. This is in contrast to MLM accuracy, which is not a continuous loss and is often stochastic (e.g., when performing dynamic masking as in RoBERTa). In \Cref{ssec:accept}, we see that PPPLs naturally separate out sets of acceptable and unacceptable sentences.

Unlike previous works \citep{Chen2017biLM, Shin2019} we use pretrained BERTs, which are \textit{open-vocabulary} (subword) bidirectional LMs. However, PPPLs are only comparable under the same subword vocabulary, which differs between e.g., BERT and RoBERTa. Normalizing with $N$ as the number of  \textit{words} mitigates this. In \Cref{ssec:pppl}, we show that word-normalized PPPLs correlate with domain adaptation, and with downstream metrics like ASR and BLEU after rescoring.

\section{Sequence-to-sequence rescoring}
\label{sec:results}
\label{sec:experimental-setup}
\label{sec:seq2seq}

Let $\mX$ denote audio features or source text tokens, and let $\mW = (\vw_1, \dotsc, \vw_{\abs{\mW}})$ denote target text tokens. For non-end-to-end ASR and MT systems, having separate acoustic/translation models $P_{\text{AM}/\text{TM}}(\mX \mid \mW)$ and language models $P_{\text{LM}}(\mW)$ is motivated by the Bayes rule decomposition used to select the best hypothesis $\hat{\mW}$ \cite{Jelinek1975, Brown1993}:
\begin{align*}
    \hat{\mW} &= \argmax_\mW [P(\mW \mid \mX)]\\
    &= \argmax_\rmW [P_{\text{AM}/\text{TM}}(\mX \mid \mW) P_{\text{LM}}(\mW)].
\end{align*}

\subsection{The log-linear model}
\label{ssec:log-linear}

End-to-end ASR and NMT use encoder-decoder architectures that are trained discriminatively. Though less principled, many still adopt a \textit{log-linear model}
\begin{align*}
    \hat{\mW} &= \argmax_\mW \left[ \log P(\mW \mid \mX) \right] \\
    &\approx \argmax_\mW \left[ \log f(\mW, \mX) + \lambda \log g(\mW) \right]
\end{align*}
with learned functions $f,g$ and a hyperparameter $\lambda$, to good effect \cite{Sutskever2014,Chan2016}. One often takes $f = P_{\text{S2S}}(\mW \mid \mX)$ as the sequence-to-sequence model and $g = P_{\text{LM}}(\mW)$ as the language model. Since the sequence-level $\argmax$ is intractable, one can do \textit{fusion}, which decomposes $f = \prod f_t$ and $g = \prod g_t$ over time \cite{Gulcehre2015}, restricting to the top $N$ intermediate candidates at each step (beam search). Instead, our work considers \textit{$N$-best rescoring}, which computes $f(\mW, \mX)$ first, still using beam search to maintain the top $N$ candidates and scores. Then, $g(\mW)$ is computed for the resulting hypotheses and interpolated with these scores, giving a new top-1 hypothesis. The sequence model is now solely responsible for ``capturing'' the best hypothesis $\hat{\mW}$ in its beam. However, there are two advantages to $N$-best rescoring, which motivate PLLs as well as our maskless finetuning approach, respectively:

\paragraph{Decoupling of scale.} Fusion requires correspondence between $f_t$ and $g_t$ at every $t$. This requires the sequence model and LM to be autoregressive and share tokenizations. In rescoring, $f = P_{\text{S2S}}$ does not require $g$ to decompose over time or to be a true probability at all, though $g$ should scale with $f$ so that $\lambda$ remains valid for all lengths $\abs{\mW}$; e.g., taking $g(\mW)$ to be a ``relevance score'' between 0 and 1 would not satisfy this property. The choice of log-linear is relevant here (\Cref{ssec:gen-model}).

\paragraph{Length-independent inference.} If $g$ is non-recurrent, then $g(\mW)$ may be computed in a single inference pass. This difference manifests with self-attentive LMs like SANLMs and Transformer-XL \cite{Dai2019}, as recently explored for $N$-best rescoring in ASR \cite{Li2019,Shin2019}.

\subsection{Experimental setup}

Further implementation and experimental details can be found in \Cref{sec:details} and our code release:

\paragraph{LMs.} We rescore sequence-to-sequence hypotheses as in \Cref{ssec:log-linear}. Each hypothesis is assigned its log probability (uni-SANLM, GPT-2) or pseudo-log-likelihood score (bi-SANLM, BERT, M-BERT, RoBERTa, XLM). We tune the LM weight $\lambda$ on the development set to minimize word error rate (WER) for ASR or maximize tokenized BLEU for NMT. We then evaluate on the test set.

\paragraph{ASR.} Our 100-best hypotheses are from an end-to-end, 5-layer BLSTMP model \citep{Shin2019} from ESPnet \cite{Watanabe2018} on the 960-hour LibriSpeech corpus \cite{Panayotov2015}. Though this baseline is not state-of-the-art, we use their lists to enable direct comparison in \Cref{table:wer-ls-adapt}.

\paragraph{NMT.} Our 100-best hypotheses are from strong Transformer baselines with BPE subwords. One was pretrained for WMT 2014 English-German \cite{Vaswani2017}; the others are state-of-the-art low-resource models we trained for five pairs from the TED Talks corpus \cite{Qi2018} and for IWSLT 2015 English-Vietnamese \cite{Cettolo2015}, which we also describe in a dedicated, concurrent work \cite{Nguyen2019}. For the low-resource models we scored tokenized hypotheses (though with HTML entities unescaped, e.g., \texttt{\&quot;}\;$\mapsto$\;\texttt{"}). Length normalization \cite{Wu2016} is applied to NMT ($\alpha=0.6$) and LM ($\alpha=1.0$) scores (\Cref{ssec:numerical}).

\begin{table}[h!]
\begin{minipage}{1.0\linewidth}
	\centering
	\footnotesize
\begin{tabu}{@{}llc@{}}
\toprule
\textbf{Corpus} & \textbf{Source \textrightarrow\ target language} & \textbf{\# pairs} \\
\midrule 
TED Talks & Galician (gl) \textrightarrow\ English (en) & 10k\\
TED Talks & Slovakian (sk) \textrightarrow\ English (en) & 61k \\
IWSLT 2015 & English (en) \textrightarrow\  Vietnamese (vi) & 133k \\
TED Talks & English (en) \textrightarrow\  German (de) & 167k \\
TED Talks & Arabic (ar) \textrightarrow\  English (en) & 214k \\
TED Talks & English (en) \textrightarrow\ Arabic (ar) & 214k \\
WMT 2014 & English (en) \textrightarrow\ German (de) & 4.5M \\
\bottomrule 
\end{tabu}
\end{minipage}
\caption{Sizes of translation datasets used in this paper.}
\label{table:nmt-sizes}
\end{table}

\subsection{Out-of-the-box (monolingual)}
\label{ssec:oob-english}

We consider BERT \cite{Devlin2019}, GPT-2 \cite{Radford2019}, and RoBERTa \cite{Liu2019}, which are trained on 17GB, 40GB, and 160GB of written text respectively. Each model comes in similarly-sized 6-layer (117M / base) and 12-layer (345M / large) versions. GPT-2 is autoregressive, while BERT and RoBERTa are MLMs. We begin by rescoring ASR outputs in \Cref{table:wer-ls-pt}:

\begin{table}[h!]
\begin{minipage}{1.0\linewidth}
	\centering
	\footnotesize
\begin{tabu}{@{}lcccc@{}}
\toprule
\multirow{2}{*}{\textbf{Model}} & \multicolumn{2}{c}{\textbf{dev}} & \multicolumn{2}{c}{\textbf{test}} \\
 & clean & other & clean & other \\
  \midrule 
baseline (100-best) & 7.17 & 19.79 & 7.26 & 20.37 \\
\midrule
GPT-2 (117M, cased) & 5.39 & 16.81 & 5.64 & 17.60 \\
BERT (base, cased) & 5.17 & 16.44 & 5.41 & 17.41 \\
RoBERTa (base, cased) & \textbf{5.03} & \textbf{16.16} & \textbf{5.25} & \textbf{17.18} \\
\midrule
GPT-2 (345M, cased) & 5.15 & 16.48 & 5.30 & 17.26 \\
BERT (large, cased) & 4.96 & 16.26 & 5.25 & 16.97 \\
RoBERTa (large, cased) & \textbf{4.75} & \textbf{15.81} & \textbf{5.05} & \textbf{16.79} \\
\midrule
\textit{oracle (100-best)} & \textit{2.85} & \textit{12.21} & \textit{2.81} & \textit{12.85} \\
\bottomrule
\end{tabu}
\end{minipage}
\caption{WERs on LibriSpeech after rescoring. Baseline lists and oracle scores are from \citet{Shin2019}.}
\label{table:wer-ls-pt}
\end{table}

As GPT-2 is trained on cased, punctuated data while the ASR model is not, we use cased MLMs and append ``\texttt{.}''\ to hypotheses to compare out-of-the-box performance. BERT outperforms its corresponding GPT-2 models despite being trained on less data. RoBERTa reduces WERs by 30\% relative on LibriSpeech \textit{test-clean} and 18\% on \textit{test-other}.

We repeat the same on English-target NMT in \Cref{table:bleu-en}. As 100-best can be worse than 4-best due to the beam search curse \cite{Yang2018, Murray2018}, we first decode both beam sizes to ensure no systematic degradation in our models. Hypothesis rescoring with BERT (base) gives up to +1.1 BLEU over our strong 100-best baselines, remaining competitive with GPT-2. Using RoBERTa (large) gives up to +1.7 BLEU over the baseline. Incidentally, we have demonstrated conclusive improvements on Transformers via LM rescoring for the first time, despite only using $N$-best lists; the most recent fusion work \cite{Stahlberg2018} only used LSTM-based models.

\begin{table}[h!]
\begin{minipage}{1.0\linewidth}
	\centering
	\footnotesize
\begin{tabu}{@{}lccc@{}}
\toprule
\multirow{2}{*}{\textbf{Model}} & \multicolumn{3}{c}{\textbf{TED Talks}} \\
 & \glTOen & \skTOen  & \arTOen \\
\midrule
\citet{Neubig2019} & 16.2 & 24.0 & -- \\
\citet{Aharoni2019} & -- & -- & 27.84 \\
our baseline (4-best) & 18.47 & 29.37 & 33.39\\
our baseline (100-best) & 18.55 & 29.20 & 33.40\\
\midrule
GPT-2 (117M, cased) & \textbf{19.24} & 30.38 & 34.41 \\
BERT (base, cased) & 19.09 & 30.27 & 34.32 \\
RoBERTa (base, cased) & 19.22 & \textbf{30.80} & \textbf{34.45} \\
\midrule
GPT-2 (345M, cased) & 19.16 & 30.76 & 34.62 \\
BERT (large, cased) & 19.30 & 30.31 & 34.47 \\
RoBERTa (large, cased) & \textbf{19.36} & \textbf{30.87} & \textbf{34.73} \\
\bottomrule
\end{tabu}
\end{minipage}
\caption{Test BLEU scores on English-target language pairs from the TED Talks corpus, after rescoring.}
\label{table:bleu-en}
\end{table}

We also consider a non-English, higher-resource target by rescoring a pre-existing WMT 2014 English-German system (trained on 4.5M sentence pairs) with German BERT (base) models\footnote{\url{https://github.com/dbmdz/german-bert}} trained on 16GB of text, similar to English BERT. From 27.3 BLEU we get +0.5, +0.3 from uncased, cased; a diminished but present effect that can be improved as in \Cref{table:bleu-en} with more pretraining, a larger model, or domain adaptation (\Cref{ssec:domain}).

\subsection{Out-of-the-box (multilingual)}

To assess the limits of our modular approach, we ask whether a shared multilingual MLM can improve translation into different target languages. We use the 100+ language M-BERT models, and the 15-language XLM models \citep{Lample2019} optionally trained with a crosslingual translation LM objective (TLM). Monolingual training was done on Wikipedia, which gives e.g., 6GB of German text; see \Cref{table:bleu-lowres}.

\begin{table}[ht]
\begin{minipage}{1.0\linewidth}
	\centering
	\footnotesize
\begin{tabu}{@{}lccc@{}}
\toprule
\multirow{2}{*}{\textbf{Model}} & \multicolumn{1}{c}{\textbf{\IWSLT}} & \multicolumn{2}{c}{\textbf{TED Talks}} \\
& \enTOvi & \enTOde & \enTOar \\
\midrule
\citet{Wang2018} & 29.09 & -- & -- \\
\citet{Aharoni2019} & -- & 23.31 & 12.95 \\
our baseline (4-best) & 31.94 & 30.50 & 13.95 \\
our baseline (100-best) & 31.84 & 30.44 & 13.94  \\
\midrule
M-BERT (base, uncased) & 32.12 & 30.48 & 13.98  \\
M-BERT (base, cased) & 32.07 & 30.45 & 13.94  \\
XLM (base*, uncased) & \textbf{32.27} & 30.61 & \textbf{14.13} \\
+ TLM objective & 32.26 & \textbf{30.62} & 14.10 \\
\midrule
de-BERT (base, uncased) & -- & \textbf{31.27} & -- \\
de-BERT (base, cased) & -- & 31.22 & -- \\
\bottomrule
\end{tabu}
\end{minipage}
\caption{Test BLEU scores for language pairs with non-English targets, after hypothesis rescoring. Base* uses 1024 hidden dimensions but only 8 heads instead.}
\label{table:bleu-lowres}
\end{table}

The 100-language M-BERT models gave no consistent improvement. The 15-language XLMs fared better, giving +0.2-0.4 BLEU, perhaps from their use of language tokens and fewer languages. Our German BERT results suggest an out-of-the-box upper bound of +0.8 BLEU, as we found with English BERT on similar resources. We expect that increasing training data and model size will boost XLM performance, as in \Cref{ssec:oob-english}.

\subsection{Domain adaptation}
\label{ssec:domain}

Out-of-the-box rescoring may be hindered by how closely our models match the downstream text. For example, our uncased multilingual models strip accents, exacerbating their domain mismatch with the cased, accented gold translation. We examine this effect in the setting of LibriSpeech, which has its own 4GB text corpus and is fully uncased and unpunctuated, unlike the cased MLMs in \Cref{ssec:oob-english}. We rescore using in-domain models in \Cref{table:wer-ls-adapt}:

\begin{table}[h!]
\begin{minipage}{1.0\linewidth}
	\centering
	\footnotesize
\begin{tabu}{@{}lcccc@{}}
\toprule
\multirow{2}{*}{\textbf{Model}} & \multicolumn{2}{c}{\textbf{dev}} & \multicolumn{2}{c}{\textbf{test}} \\
 & clean & other & clean & other \\
  \midrule 
baseline (100-best) & 7.17 & 19.79 & 7.26 & 20.37 \\
\midrule
uni-SANLM & 6.08 & 17.32 & 6.11 & 18.13 \\
bi-SANLM & 5.52 & 16.61 & 5.65 & 17.44 \\
BERT (base, Libri.\ only) & \textbf{4.63} & \textbf{15.56} & \textbf{4.79} & \textbf{16.50} \\
\midrule
BERT (base, cased) & 5.17 & 16.44 & 5.41 & 17.41 \\
BERT (base, uncased) & 5.02 & 16.07 & 5.14 & 16.97 \\
+ adaptation, 380k steps & \textbf{4.37} & \textbf{15.17} & \textbf{4.58} & \textbf{15.96} \\
\midrule
\textit{oracle (100-best)} & \textit{2.85} & \textit{12.21} & \textit{2.81} & \textit{12.85} \\
\bottomrule
\end{tabu}
\end{minipage}
\caption{WERs on LibriSpeech after hypothesis rescoring. Baseline, SANLM, and oracle numbers are from \citet{Shin2019}.}
\label{table:wer-ls-adapt}
\end{table}

Using a BERT model trained only on the text corpus outperforms RoBERTa (\Cref{table:wer-ls-pt}) which is trained on far more data, underscoring the tradeoff between in-domain modeling and out-of-the-box integration. Even minor differences like casing gives +0.3-0.4 WER at test time. In \Cref{ssec:numerical} we see that these domain shifts can be visibly observed from the positionwise scores $\log P_{\text{MLM}}(\vw_t \mid \mW_{\backslash t})$.

The best results (``adaptation'') still come from adapting a pretrained model to the target corpus. We proceed as in BERT, i.e., performing MLM on sequences of concatenated sentences (more details in \Cref{sec:details}). In contrast, the 3-layer SANLMs \citep{Shin2019} do per-utterance training, which is slower but may reduce mismatch even further.

Finally, we show in \Cref{ssec:pppl} that even before evaluating WER or BLEU, one can anticipate improvements in the downstream metric by looking at improvements in word-normalized PPPL on the target corpus. The domain-adapted MLM has lower PPPLs than the pretrained models, and RoBERTa has lower PPPLs than BERT.

\subsection{Finetuning without masking}

We finetune BERT to produce scores without \texttt{[MASK]} tokens. For LibriSpeech we take the normalized text corpus and keep sentences with length $\abs{\mW} \le$ 384, score them with our adapted BERT (base), then do sentence-level regression (\Cref{ssec:maskless-scoring}). We train using Adam with a learning rate of $10^{-5}$ for 10 epochs (\Cref{table:wer-ls-kd}):

\begin{table}[h!]
\begin{minipage}{1.0\linewidth}
	\centering
	\footnotesize
\begin{tabu}{@{}lcc@{}}
\toprule
\multirow{2}{*}{\textbf{Model}} & \multicolumn{2}{c}{\textbf{dev}} \\
& clean & other \\
\midrule
baseline (100-best) & 7.17 & 19.79 \\
\midrule
GPT-2 (117M, cased) & 5.39 & 16.81 \\
BERT (base, uncased, adapted) & 4.37 & 15.17 \\
+ no masking & 5.79 & 18.07 \\
+ sentence-level finetuning & 4.61 & 15.53 \\
\bottomrule
\end{tabu}
\end{minipage}
\caption{WERs on LibriSpeech upon rescoring, showing the effects of single-copy, maskless scoring.}
\label{table:wer-ls-kd}
\end{table}

Sentence-level finetuning degrades performance by +0.2-0.4 WER, leaving room for future improvement. This still outperforms GPT-2 (117M, cased), though this gap may be closed by adaptation. For now, maskless finetuning could be reserved for cases where only a masked language model is available, or when latency is essential.

Remarkably, we found that out-of-the-box scoring without \texttt{[MASK]} still significantly improves the baseline. This is likely from the 20\% of the time BERT does not train on \texttt{[MASK]}, but instead inputs a random word or the same word \cite{Devlin2019}. Future work could explore finetuning to positionwise distributions, as in word-level knowledge distillation \cite{Kim2016}, for which our results are a na\"{i}ve performance bound.
\section{Analysis}
\label{sec:analysis}
\label{ssec:fluency}

We recall the log-linear model from \Cref{ssec:log-linear}:
\begin{align*}
    \hat{\mW} &\approx \argmax_\mW \left[ \log f(\mW, \mX) + \lambda \log g(\mW) \right]
\end{align*}
Although end-to-end models $f = P_{\text{S2S}}(\mW | \mX)$ predict $\mW$ directly from $\mX$, interpolation with the unconditional $g = P_{\text{LM}}(\mW)$ remains helpful \citep{Toshniwal2018}. One explanation comes from \keyword{cold} and \keyword{simple fusion} \citep{Sriram2018, Stahlberg2018}, which further improve on shallow fusion (\Cref{ssec:log-linear}) by learning $g(\mW)$ first. They argue $g$ expresses \keyword{fluency}; fixing $g$ early allows $f(\mW, \mX)$ to focus its capacity on \keyword{adequacy} in encoding the source, and thus specializing the two models. With this perspective in mind, we compare $\log P_{\text{LM}}$ and PLL as candidates for $\log g$.

\subsection{Relative linguistic acceptability}
\label{ssec:accept}

\newcommand\rotation{30} 
\begin{table*}[ht]
\begin{minipage}{1.0\linewidth}
	\centering
	\footnotesize
	\addtolength{\tabcolsep}{-3.8pt}

\begin{tabu}{@{}ll@{\hskip 8pt}llllllllllll@{\hskip 8pt}lll@{}}
\multicolumn{1}{p{2.5ex}}{\rotatebox{0}{\textbf{Model (cased)}}} &
\multicolumn{1}{p{2.5ex}}{\rotatebox{\rotation}{\textbf{Overall}}} &
\multicolumn{1}{p{2.5ex}}{\rotatebox{\rotation}{\textsc{Ana.\ agr}}} &
\multicolumn{1}{p{2.5ex}}{\rotatebox{\rotation}{\textsc{Arg.\ str}}} &
\multicolumn{1}{p{2.5ex}}{\rotatebox{\rotation}{\textsc{Binding}}} &
\multicolumn{1}{p{2.5ex}}{\rotatebox{\rotation}{\textsc{Ctrl.\ rais.}}} &
\multicolumn{1}{p{2.5ex}}{\rotatebox{\rotation}{\textsc{D-n agr}}} &
\multicolumn{1}{p{2.5ex}}{\rotatebox{\rotation}{\textsc{Ellipsis}}} &
\multicolumn{1}{p{2.5ex}}{\rotatebox{\rotation}{\textsc{Filler\ gap}}} &
\multicolumn{1}{p{2.5ex}}{\rotatebox{\rotation}{\textsc{Irregular}}} &
\multicolumn{1}{p{2.5ex}}{\rotatebox{\rotation}{\textsc{Island}}} &
\multicolumn{1}{p{2.5ex}}{\rotatebox{\rotation}{\textsc{NPI}}} &
\multicolumn{1}{p{2.5ex}}{\rotatebox{\rotation}{\textsc{Quantifiers}}} &
\multicolumn{1}{p{2.5ex}}{\rotatebox{\rotation}{\textsc{S-v agr}}} &
\multicolumn{1}{p{2.5ex}}{\rotatebox{\rotation}{Unacc. PPPL}} &
\multicolumn{1}{p{2.5ex}}{\rotatebox{\rotation}{Acc. PPPL}} &
\multicolumn{1}{p{2.5ex}}{\rotatebox{\rotation}{Ratio}} \\
\toprule 
GPT-2 (345M) & 82.6 & \textbf{99.4} & 83.4 & 77.8 & 83.0 & 96.3 & 86.3 & 81.3 & 94.9 & 71.7 & 74.7 & \textbf{74.1} & 88.3 & -- & -- & -- \\
\midrule
BERT (base) & 84.2* & 97.0 & 80.0 & \textbf{82.3}* & 79.6 & \textbf{97.6}* & 89.4* & 83.1* & 96.5* & 73.6* & 84.7* & 71.2 & \textbf{92.4}*   & 111.2 & 59.2  & 1.88 \\
BERT (large) & 84.8* & 97.2 & 80.7 & 82.0* & 82.7 & \textbf{97.6}* & 86.4 & 84.3* & 92.8 & 77.0* & 83.4* & 72.8 & 91.9*  & 128.1 & 63.6 & 2.02 \\
RoBERTa (base) & 85.4* & 97.3 & 83.5 & 77.8 & 81.9 & 97.0 & \textbf{91.4}* & \textbf{90.1}* & 96.2* & 80.7* & 81.0* & 69.8 & 91.9*  &  213.5 &  87.9 & 2.42 \\
RoBERTa (large) & \textbf{86.5}* & 97.8 &\textbf{84.6}* & 79.1* & \textbf{84.1}* & 96.8 & 90.8* & 88.9* & \textbf{96.8}* & \textbf{83.4}* & \textbf{85.5}* & 70.2 & 91.4*    & 194.0 & 77.9 & 2.49 \\
\midrule
\textit{Human} & \textit{88.6} & \textit{97.5} & \textit{90.0} & \textit{87.3} & \textit{83.9} & \textit{92.2} & \textit{85.0} & \textit{86.9} & \textit{97.0} & \textit{84.9} & \textit{88.1} & \textit{86.6} & \textit{90.9} & -- & -- & -- \\
\bottomrule

\end{tabu}
\end{minipage}
\caption{Unsupervised performance (forced choice accuracy) on BLiMP using log probabilities (GPT-2) or PLLs. Human scores from \citet{Warstadt2019}. Values with * denote improvements over GPT-2 of $\ge$1\% absolute.}
\label{table:accuracy-blimp}
\end{table*}

In this work we interpret fluency as linguistic \keyword{acceptability} \citep{Chomsky1957}; informally, the syntactic and semantic validity of a sentence according to human judgments \citep{Schutze1996}. Its graded form is well-proxied by neural language model scores ($\log P_{\text{LM}}$) once length and lexical frequency are accounted for \cite{Lau2017}. This can be seen in a controlled setting using \textit{minimal pairs} and GPT-2 (345M) scores:
\begin{align*}
    \text{\checkmark}\quad& \text{\textit{Raymond is selling this sketch.}} &-40.0,\\
     \quad& \text{\textit{Raymond is selling this sketches.}} &-45.2.
\end{align*}
This example is from the Benchmark of Linguistic Minimal Pairs (BLiMP) \citep{Warstadt2019}, a challenge set of 67k pairs which isolate contrasts in syntax, morphology, and semantics (in this example, determiner-noun agreement).  While its predecessor, the Corpus of Linguistic Acceptability (CoLA), has a training set and asks to label sentences as ``acceptable'' or not in isolation \citep{Warstadt2019cola}, BLiMP provides an unsupervised setting: language models are evaluated on how often they give the acceptable sentence a higher (i.e., less negative) score. This is equivalent to 2-best rescoring without sequence model scores ($\log f = 0$). Since most minimal pairs only differ by a single word, the effect of length on log probabilities and PLLs (discussed in \Cref{ssec:numerical}) is mitigated.

We compute PLLs on the sentences of each pair using cased BERT and RoBERTa, then choose the sentence with the highest score. Our results are in \Cref{table:accuracy-blimp}. Despite using less than half the data and a third of the capacity, BERT (base) already outperforms the previous state of the art (GPT-2) by 1.6\% absolute, increasing to 3.9\% with RoBERTa (large). There are 4 of 12 categories where all four PLLs outperform log probabilities by $\ge$1\% absolute (values marked by *), and 7 where three or more PLLs outperform by this margin. Interestingly, PLLs do consistently worse on quantifiers, though all are relatively bad against the human baseline. The ratio of token-level PPPLs between unacceptable and acceptable sentences overall increases with performance, separating the two sentence sets.

RoBERTa improves by around 10\% on filler-gap dependencies, island effects, and negative polarity items (NPIs), largely closing the human gap. This suggests that the difficulty of these BLiMP categories was due to $P_{\text{LM}}$ decomposing autoregressively, and not intrinsic to unsupervised language model training, as the original results may suggest \citep{Warstadt2019}. For some intuition, we include examples in \Cref{table:fluency}. In the subject-verb agreement example, BERT sees \textit{The pamphlets} and \textit{resembled those photographs} when scoring \textit{have} vs.\ \textit{has}, whereas GPT-2 only sees \textit{The pamphlets}, which may not be enough to counter the misleading adjacent entity \textit{Winston Churchill} at scoring time.

\subsection{Interpolation with direct models}

\begin{table*}
    \centering
    \scriptsize
    \begin{tabular}{@{}llp{280pt}@{}}
        \toprule
        \footnotesize \textbf{System} & \footnotesize \textbf{Model} & \footnotesize \textbf{Output sentence} \\
        
        \midrule

        \multirow{2}{*}{\footnotesize BLiMP (S-V agreement)} & BERT & The pamphlets about Winston Churchill \hl{\textbf{have}} resembled those photographs. \\
        & GPT-2 & The pamphlets about Winston Churchill \hl{\textbf{has}} resembled those photographs. \\
        
        \midrule
        
        \multirow{2}{*}{\footnotesize BLiMP (island)}  & BERT & Who does Amanda find \hl{\textbf{while thinking about}} Lucille? \\
        & GPT-2 & Who does Amanda find Lucille \hl{\textbf{while thinking about}}? \\
        
        \midrule

        \multirow{4}{*}{\footnotesize LibriSpeech (dev-other)} & Baseline & \hl{\textbf{clasping truth and jail ya in}} the mouth of the student is that building up or tearing down \\
        & GPT-2 & \hl{\textbf{class in truth and jail ya in}} the mouth of the student is that building up or tearing down\\
        & BERT (adapted) & \hl{\textbf{clasping truth in jail gagging}} the mouth of the student is that building up or tearing down \\
        & Target & \hl{\textbf{clapping truth into jail gagging}} the mouth of the student is that building up or tearing down \\

        \midrule

        \multirow{5}{*}{\footnotesize \glTOen\ (test)} & Source (gl) & Traballaba de asesora cient\'{i}fica na \hl{\textbf{ACLU , a Uni\'{o}n polas Liberdades Civ\'{i}s}} . \\
        &  Baseline & I worked on a scientific status on the \hl{\textbf{ACL, the Union by the Union Sivities}} . \\
        &  GPT-2 & I worked on a scientific status on the \hl{\textbf{ACL, the Union by the Union by the Union Civities}} . \\
        &  BERT & I worked on a scientific status on the \hl{\textbf{ACL, the Union by the Union of LiberCivities}} . \\
        & Target (en) & I was working at the \hl{\textbf{ACLU}} as the organization 's science advisor . \\

        \bottomrule
    \end{tabular}
        \caption{Examples of different top-1 hypotheses after ranking the minimal pairs or rescoring hypotheses from 4-best models, with differences highlighted. GPT-2 and BERT both promote fluency, but GPT-2's left-to-right biased scores appear to cause it to overweigh common word sequences at the expense of adequacy.}
    \label{table:fluency}
\end{table*}

We observed that $\log g = \text{PLL}(\mW)$ is not unduly affected by unconditional token frequencies; this mitigates degradation in adequacy upon interpolation with $P_{\text{S2S}}$. Consider a two-word proper noun, e.g., $\mW = \text{``San Francisco''}$:
\begin{align*}
    &\log P_{\text{LM}}(\mW)\\
    &= \log P_{\text{LM}}(\text{San}) + \log P_{\text{LM}}(\text{Francisco} \mid \text{San})\\
    &\ll \log P_{\text{MLM}}(\text{San} \mid \text{Francisco})\\
    &\quad \quad + \log P_{\text{MLM}}(\text{Francisco} \mid \text{San})\\
    &= \text{PLL}(\mW).
\end{align*}
It is a highly-fluent but low-probability bigram and thus gets penalized by $\log P_{\text{LM}}(\mW)$. Informally, $\text{PLL}(\mW)$ expresses how likely each token is given other tokens (self-consistency), while $\log P_{\text{LM}}(\mW)$ expresses the unconditional probability of a sentence, beginning with the costly unconditional term $P_{\text{LM}}(\text{San})$. We see this in practice when we take LM to be GPT-2 (345M) and MLM to be RoBERTa (large). Substituting in the actual scores:
\begin{align*}
    \log P_{\text{GPT-2}}(\mW) &= -8.693 \\
    &= (-7.749) + (-0.944)\\
    &\ll (-0.006) + (-1.000) \\
    &= -1.006 = \text{PLL}_{\text{RoBERTa}}(\mW).
\end{align*}
Both give similar probabilities $P(\text{Francisco} \mid \text{San})$ $\approx$ $e^{-1.0}$ $\approx$ 37\%, but differ in the first summand.

We examine the interplay of this bias with our sequence models, in cases where the baseline, GPT-2, and BERT gave different top-1 hypotheses (\Cref{table:fluency}). In our examples, GPT-2 restores fluency using common and repeated words, at the cost of adequacy:
\begin{align*}
    &\text{\textit{clasping truth and}} \mapsto \text{\textit{class in truth and}},\\
    &\text{\textit{Union by the Union Sivities}} \mapsto\\
    &\quad \text{\textit{Union by the Union by the Union Civities}}.
\end{align*}
One can view these as exacerbations of the rare word problem due to overconfident logits \citep{Nguyen2018}, and of over-translation \cite{Tu2016}. Meanwhile, BERT rewards self-consistency, which lets rarer but still-fluent words with better acoustic or translation scores to persist:
\begin{align*}
    &\text{\textit{clasping truth and}} \mapsto \text{\textit{clasping truth in}},\\
    &\text{\textit{Union by the Union Sivities}} \mapsto\\
    &\quad \text{\textit{Union by the Union of LiberCivities}},
\end{align*}
which preserves the \textit{p} sound in the ground truth (\textit{clapping}) for ASR, and promotes the more globally-fluent \textit{Union by the Union \textbf{of} LiberCivities}. We also see the under-translation (i.e., omission) of \textit{Liber} being corrected, without being discouraged by the rare sequence \textit{LiberCivities}.

Given the differences between PLLs and log probabilities, we explore whether ensembling both improves performance in \Cref{ssec:combo}. Similar to the largely-dominant results of MLMs on BLiMP over GPT-2 (\Cref{ssec:accept}), we find that as the MLM gets stronger, adding GPT-2 scores has negligible effect, suggesting that their roles overlap.

\subsection{Numerical properties of PLL}
\label{ssec:numerical}

\begin{figure}[ht!]
\centering
\includegraphics[width=0.95\linewidth,trim={0 0.35cm 0 0.25cm},clip]{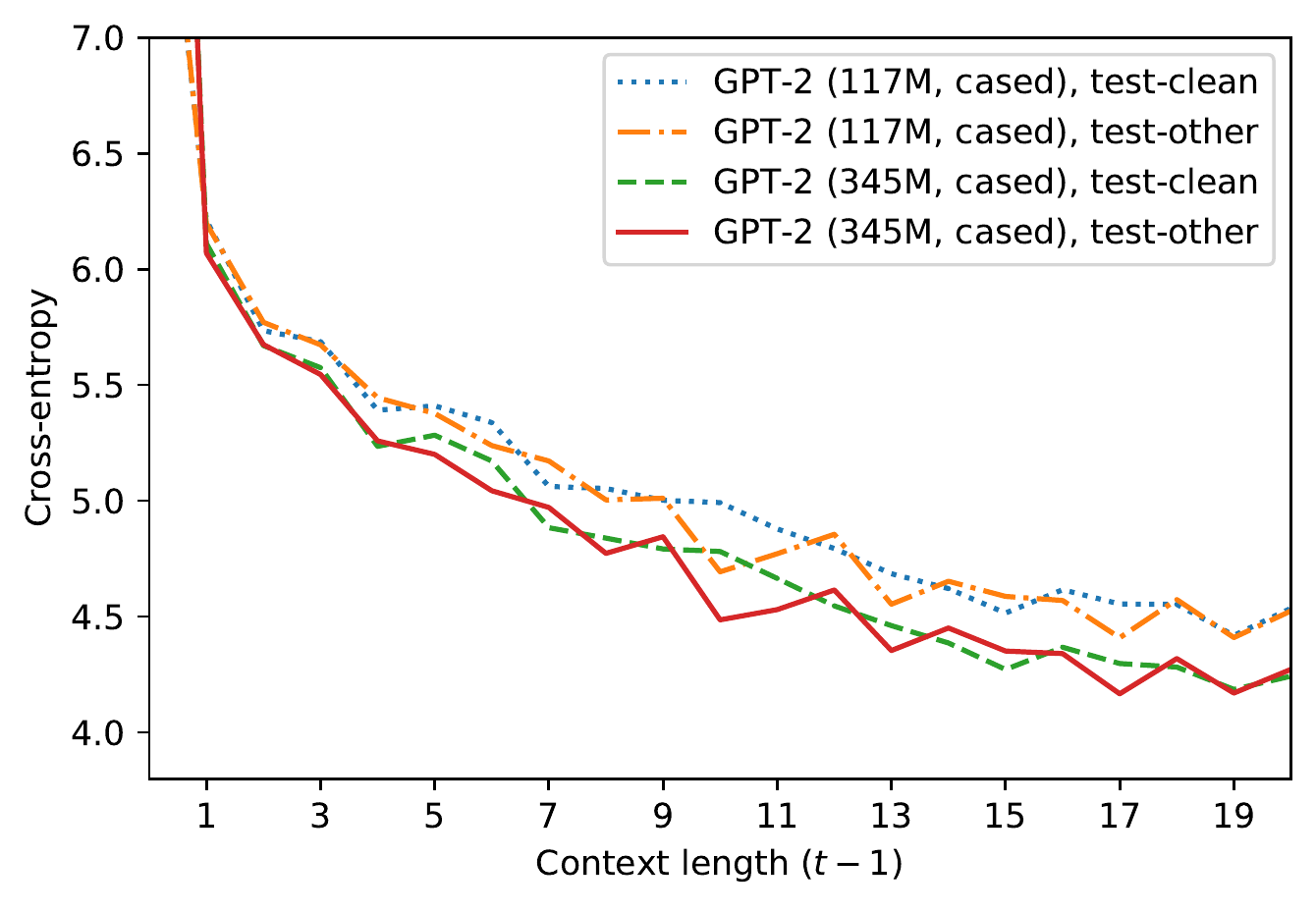}
\caption{Cross-entropy (natural base) of $\rvw_t \mid \mW_{< t}$ versus context length ($t-1$) from GPT-2 models, averaged over LibriSpeech's test utterances.}
\label{fig:gpt2-ce}
\end{figure}

\begin{figure}[t!]
\centering
\includegraphics[width=0.95\linewidth,trim={0 0.35cm 0 0.25cm},clip]{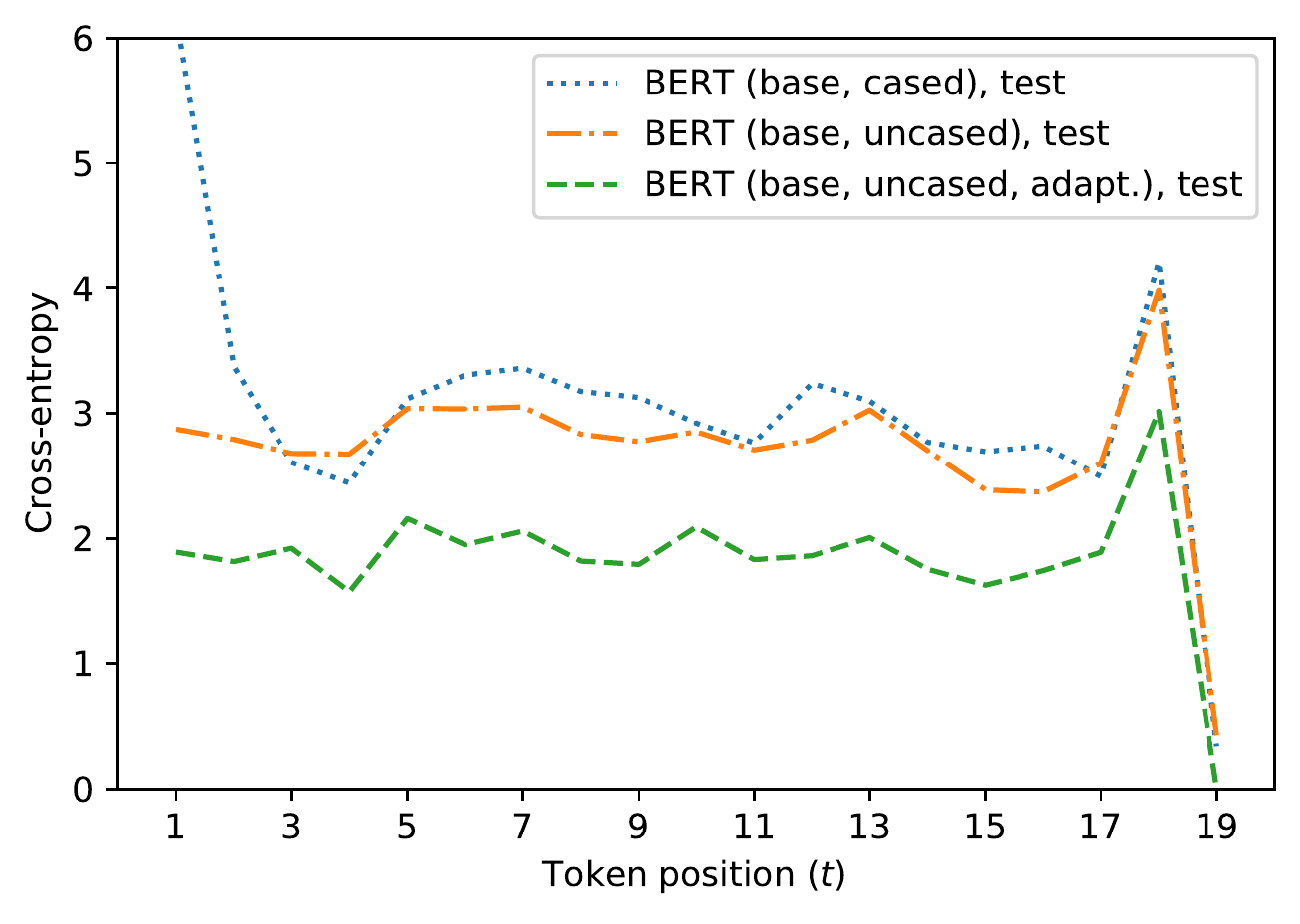}
\caption{Cross-entropy (natural base) of $\rvw_t \mid \mW_{\backslash t}$ versus $t$ from BERT, averaged over LibriSpeech's 189 test utterances of length $\abs{\mW} = 19$ (including ``\texttt{.}'').}
\label{fig:bert-ce}
\end{figure}

PLL's numerical properties make it an ideal foundation for future ranking or scoring schemes. For example, given fixed $\abs{\mW}$ one expects $-\log P_{\text{MLM}}(\vw_t \mid \mW_{\backslash t})$ to be in the same range for all $t$. Meanwhile $-\log P_{\text{LM}}(\vw_t \mid \mW_{< t})$ decreases as $t \to \abs{\mW}$, the rate of which was studied in recurrent language models \citep{Takahashi2018}. We validate this with GPT-2 (\Cref{fig:gpt2-ce}) and BERT (\Cref{fig:bert-ce}). In particular, we see the outsized cost of the unconditional first unigram in \Cref{fig:gpt2-ce}. This also explains why bi-SANLM was more robust than uni-SANLM at shorter and earlier positions \citep{Shin2019}; the difference is intrinsic to log probabilities versus PLLs, and is not due to model or data size.


\Cref{fig:bert-ce} also shows that domain adaptation (\Cref{ssec:domain}) affects PLL's positionwise cross-entropies. Cased BERT spikes at position 1, as it observes a lowercase word where a capitalized word is expected. All MLMs spike at the final token of an utterance, before our appended period ``\texttt{.}''. Terminal words are difficult to predict in general, but here more so as the BERT+LibriSpeech text corpora and the LibriSpeech test set are mismatched;  the latter's ground-truth utterances were segmented by voice activity and not punctuation \citep{Panayotov2015}. Otherwise, the averaged cross-entropies are flat. This, plus our success on BLiMP, suggest positionwise scores as a way of detecting ``disfluencies'' (at least, those in the form of domain mismatches) by observing spikes in cross-entropy; with $\log P_{\text{LM}}$, spikes are confounded by the curve in \Cref{fig:gpt2-ce}.

In \Cref{ssec:pppl}, we plot sentence-level PLLs versus $\abs{\mW}$ and observe linearity as $\abs{\mW} \to \infty$, with spikes from the last word and lowercase first word smoothing out. This behavior motivates our choice of $\alpha=1.0$ when applying the Google NMT-style length penalty \citep{Wu2016} to PLLs, which corresponds to the asymptotically-linear $\text{LP}_{\text{MLM}} = (5 + \abs{\mW}) / (5+1)$. In contrast, autoregressive scores like $P_{\text{LM}}(\mW)$ integrate over the inverse power-law curve in \Cref{fig:gpt2-ce}. We speculate that this explains the effectiveness of their hyperparameter $\alpha = 0.6$, widely used in NMT baselines like ours, as there exists $C$ such that
\begin{align*}
    \text{LP}_{\text{S2S}}(\mW) = \frac{(5 + \abs{\mW})^{0.6}}{(5+1)^{0.6}}\!\approx\! \int_0^{\abs{\mW}}\!\!\!\!\frac{C}{(5+x)^{0.4}}\;\textrm{d}x.
\end{align*}

\section{Related work}
\label{sec:related-work}

Our work extends the closest previous works \citep{Wang2019, Shin2019} with regards to experiments and tasks, as outlined in \Cref{ssec:pseudolikelihood}. Furthermore, neither work considers the inference cost of masked rescoring, which we address with our maskless scoring approach, or analyze PLL's numerical properties.

\paragraph{Future context.} Log probabilities conditioned on past and future context have been used in MT \citep{Finch2009, Xiong2011} and perennially in ASR \citep{Shi2013, Arisoy2015, Chen2017biLM} to positive effect. However, these are not ``deep bidirectional'' as they model interactions between $\mW_{<t}$ and $\mW_{>t}$ via the forward and backward context vectors, while MLMs model all pairwise interactions $\vw_s$ and $\vw_{s'}$ via dot-product attention (compare ELMo versus BERT). Their PLLs would have different properties from ours (e.g., their cross-entropies in \Cref{fig:bert-ce} may be convex instead of flat).

\paragraph{Discriminative language modeling.} Previous works \citep{Roark2004, Huang2018} have explored training language models that directly optimize for a downstream metric (WER, BLEU). While we also eschew using log probabilities from conventional LMs, our approach remains generative. Log probabilities model the joint distribution; PLL does so as well, albeit implicitly (\Cref{ssec:gen-model}). PLL's summands (conditional probabilities) remain accessible for Gibbs sampling and are not tailored to any metric. The two approaches are complementary; for example, one could use PLL as a ``prior'' or regularizer for scores given by discriminatively-finetuned BERT models in tasks like passage re-ranking \citep{Nogueira2019}.

\paragraph{Language model integration.} Beyond finetuning pretrained LMs and MLMs, monolingual pretraining has also improved NMT performance \citep{Ramachandran2017, Lample2019}. However, modular integration of language representation models remains prevalent for various pragmatic reasons, similar to fusion in ASR. Contemporary examples are the use of finetuned BERT scores in a question-answering pipeline \citep{Nogueira2019}, or ``as-is'' cosine similarity scores from BERT to evaluate generated text \citep{Zhang2019}. For example, one might have no pretrained multilingual LMs for decoder initialization or fusion, as such models are difficult to train \cite{Ragni2016}. However, one may have an M-BERT or XLM for the target language/domain. Finally, $N$-best rescoring and pretraining are not mutually exclusive, though pretraining may already go partway to improve fluency.
\section{Conclusion}
\label{sec:conclusion}

We studied scoring with MLM \textit{pseudo-log-likelihood scores} in a variety of settings. We showed the effectiveness of $N$-best rescoring with PLLs from pretrained MLMs in modern sequence-to-sequence models, for both ASR and low- to medium-resource NMT. We found rescoring with PLLs can match or outperform comparable scores from large unidirectional language models (GPT-2).  We attributed this to PLL's promotion of fluency via self-consistency, as demonstrated by improvement on unsupervised acceptability judgements and by qualitative analysis. We examined the numerical properties of PLLs, proposed maskless scoring for speed, and proposed \textit{pseudo-perplexities} for intrinsic evaluation of MLMs, releasing a codebase implementing our work. Future work could find additional modular uses of MLMs, simplify maskless PLL computations, and use PLLs to devise better sentence- or document-level scoring metrics.

\section*{Acknowledgments}

We thank Phillip Keung and Chris Varano for their thoughtful suggestions on this work.

\bibliography{paper}
\bibliographystyle{acl_natbib}

\newpage
\appendix
\section{Experiment details}
\label{sec:details}

\subsection{Language models}

\paragraph{Implementation.} English BERT, M-BERT, GPT-2, and RoBERTa models were served, adapted, and finetuned via the GluonNLP toolkit \citep{Guo2019}. German BERT and XLM models were served via HuggingFace's Transformers toolkit \citep{Wolf2019}. We release a reference implementation (a language model scoring package) for our work at \url{https://github.com/awslabs/mlm-scoring}.

\paragraph{Training.} When adapting to a corpus we continue the training scheme for BERT, i.e., MLM + next-sentence prediction \citep{Devlin2019}, on the new dataset only, until the training loss converges. We still perform warmup at adaptation time (ratio of 0.01), but continue to use batches of 256 sequences of contiguous sentences, each with length up to 512.

\paragraph{Scoring.} For BERT, M-BERT, and RoBERTa we prepend and append \texttt{[CLS]}, \texttt{[SEP]} tokens. For GPT-2 we prepend and append \texttt{<|endoftext|>}, the default tokens for unconditional generation, as we found this outperformed other initial conditions (e.g., a preceding ``\texttt{.}''). For XLM we prepend and append \texttt{</s>} (prepending \texttt{<s>} is more proper, but this is due to a bug in HuggingFace Transformer's \texttt{XLMTokenizer} that we will fix; changes in results should be negligible). When computing (pseudo-)perplexity (\Cref{ssec:pppl-intro}), these special tokens' conditional probabilities are not included, nor are they counted for token or word counts during length normalization.

\paragraph{$N$-best rescoring.} We follow the log-linear model in \Cref{ssec:log-linear} with its hyperparameter $\lambda$, i.e., weighted addition of (M)LM scores with sequence-to-sequence scores. When interpolating MLMs with GPT-2 there is also a hyperparamter $\gamma$ (\Cref{ssec:combo}). We do grid search on $(\lambda, \gamma)$ with increments (0.05, 0.1) for the best weights on the development set for downstream WER or BLEU, then evaluate on the corresponding test set. In the case of ties, we choose the largest $\lambda$, $\gamma$.

\subsection{Automatic speech recognition}

We use the LibriSpeech corpus \citep{Panayotov2015} for our experiments. To adapt BERT we use the provided 800M-word text-only data, processed using Kaldi to match the normalized, downloadable corpus\footnote{\url{https://www.openslr.org/resources/11/librispeech-lm-norm.txt.gz}} but \keyword{with sentences in their original order} (instead of alphabetically as in Kaldi's recipe), to match the long-context training regime of our language models. Our LibriSpeech-only BERT (base) model was trained on this corpus using GluonNLP's recipe, for 1.5M steps.

We take pre-existing 100-best lists shared via e-mail communication \citep{Shin2019}, which were produced by ESPnet \citep{Watanabe2018} on LibriSpeech's dev and test sets. The ESPnet model was the sequence-to-sequence BLSTMP model in the \textit{librispeech/asr1} recipe, except with 5 layers and a beam size of 100.

For speech corpora, to alleviate some of the domain shift from BERT's original written corpora, we appended ``\texttt{.}''\ at the end of utterances during adaptation, and appended ``\texttt{.}''\ to all hypotheses before subword tokenization, masking, and token/word counting.

\subsection{Neural machine translation}

Our pretrained model\footnote{\url{http://apache-mxnet.s3-accelerate.dualstack.amazonaws.com/gluon/models/transformer_en_de_512_WMT2014-e25287c5.zip}} is the base Transformer on WMT 2014 English-German \citep{Vaswani2017} trained using GluonNLP's \url{scripts/machine_translation}. Evaluation and $N$-best rescoring was on the 3003-sentence test set via \texttt{--full --bleu 13a --beam\_size 100}.

We consider 5 low-resource directions from the TED Talks dataset \citep{Qi2018}: Arabic (ar), Galician (gl), and Slovak (sk) to English; and English to Arabic, German (de), languages which were considered in \citet{Aharoni2019}. We also include a more popular benchmark, English to Vietnamese (vi) from the \IWSLT\ evaluation campaign\footnote{\url{https://nlp.stanford.edu/projects/nmt/}} \citep{Cettolo2015}. These give a breadth of English-source and English-target pairs and include a right-to-left language; more importantly, the three non-English targets are covered by the 15-language XLMs \cite{Lample2019}.

Our models are also described as baselines in a dedicated work \citep{Nguyen2019}. They are base Transformers with 6 layers, 8 heads, an 8k BPE vocabulary, and dropout of 0.3, except for \glTOen\ where we use 4 layers, 4 heads, 3k BPE, and a dropout of 0.4 due to its significantly smaller size. We use a warmup of 8k steps and the default hyperparameters \citep{Vaswani2017}. We apply GNMT length normalization \cite{Wu2016} with $\alpha = 0.6$ to the sequence-to-sequence log probabilities, and $\alpha = 1.0$ to the PLLs (motivation is given in \Cref{ssec:numerical}), with respect to their chosen tokenization's lengths. We compute tokenized BLEU via  \textit{multi-bleu.perl} from Moses\footnote{\url{https://statmt.org}} to compare with past works on these datasets.

\section{BERT as a generative model}
\label{ssec:gen-model}

In their published version \citep{Wang2019}, the authors claimed that BERT is a Markov random field language model (MRF-LM) where $\{\rvw_t\}_{t=1}^{\abs{\mW}}$ are categorical random variables (over the vocabulary) in a fully-connected graph $\gG$. They define a \textit{potential} over cliques of $\gG$ such that all partial-graph potentials are $\exp(0) = 1$ and the full-graph potential is $\exp \sum_{t=1}^{\abs{\mW}} \log \phi_t(\gG)$, where $\log \phi_t(\gG)$ is the logit corresponding to $\log P_{\text{MLM}}(\vw_t \mid \mW_{\backslash t})$ (although in their formulation, one could include the softmax into the feature function $f_\theta$ and take $\log \phi_t(\gG) = \text{PLL}(\gG)$ exactly).

Abusing notation, we write $\mW$ interchangeably with its graph $\gG$. An MRF defined this way would give the joint distribution:
\begin{align*}
P_{\text{MLM}}(\mW) = \frac{1}{Z} \prod_{t=1}^{\abs{\mW}} \phi_t(\mW) = \frac{1}{Z} \exp \text{PLL}(\mW),
\end{align*}
where $Z$ is the partition function
\begin{align*}
Z = \sum_{\mW' \in \sS} \prod_{t=1}^{\abs{\mW'}} \phi_t(\mW') = \sum_{\mW' \in \sS}\!\!\exp \text{PLL}(\mW'),
\end{align*}
making this a valid distribution by normalizing over all sequences of the same length $\abs{\mW}$, the set denoted by $\sS$.

One then hopes to say that $\log P_{\text{MLM}}(\vw_t \mid \mW_{\backslash t})$ is the conditional distribution of this MRF. However, their erratum\footnote{\href{https://sites.google.com/site/deepernn/home/blog/amistakeinwangchoberthasamouthanditmustspeakbertasamarkovrandomfieldlanguagemodel}{``BERT has a Mouth and must Speak, but it is not an MRF''} from \url{https://sites.google.com/site/deepernn/home/blog}} notes this is not the case, as $\rvw_t$ would be affected by other log potentials as well.

In practice, one could instead \textit{a priori} make the modeling assumption
\begin{align*}
    g(\mW) = P_{\text{MLM}}(\mW) := \frac{1}{Z} \exp \text{PLL}(\mW),
\end{align*}
as done in the work on bi-RNNLMs \citep{Chen2017biLM}. They \textit{choose} to model the distribution of sentences as a product-of-experts $\rvw_t \mid \mW_{\backslash t}$, whose parameters are shared via the underlying bi-RNN.

Suppose one had access to this ``normalized MLM probability''. In the log-linear setting (\Cref{ssec:log-linear}), we get
\begin{align*}
\log\;&P_{\text{S2S}}(\mW \mid \mX) + \lambda \log P_{\text{MLM}}(\mW)\\
&= \dotsb + \lambda \log \left(\frac{1}{Z} \exp \text{PLL}(\mW)\right)\\
&= \dotsb + \lambda\;\text{PLL}(\mW) - \lambda \log Z.
\end{align*}
For fixed $\lambda$ and $Z$ (which is intrinsic to the MLM), we see that $\lambda \log Z$ does not affect rank-ordering when taking $\argmax$ to get the best hypothesis $\hat{\mW}$. Hence, the heuristic interpolation enacted by $\lambda$ is ``the same'' for normalized $\log P_{\text{LM}}$, unnormalized PLL, and our hypothetical $\log P_{\text{MLM}}$. The remaining issue is whether $\lambda$ has the same effect for all lengths $\abs{\mW}$, which one mitigates by applying the correct length penalties to $f$ and $g$ (\Cref{ssec:numerical}).

\section{Pseudo-perplexity and rescoring}
\label{ssec:pppl}

We briefly examine the relationship between PPPL (\Cref{ssec:pppl-intro}) and metrics post-rescoring. We plot negative PLLs versus $\abs{\mW}$ and observe linearity, helping justify our simple average over length:

\begin{figure}[ht!]
\centering
\includegraphics[width=0.95\columnwidth]{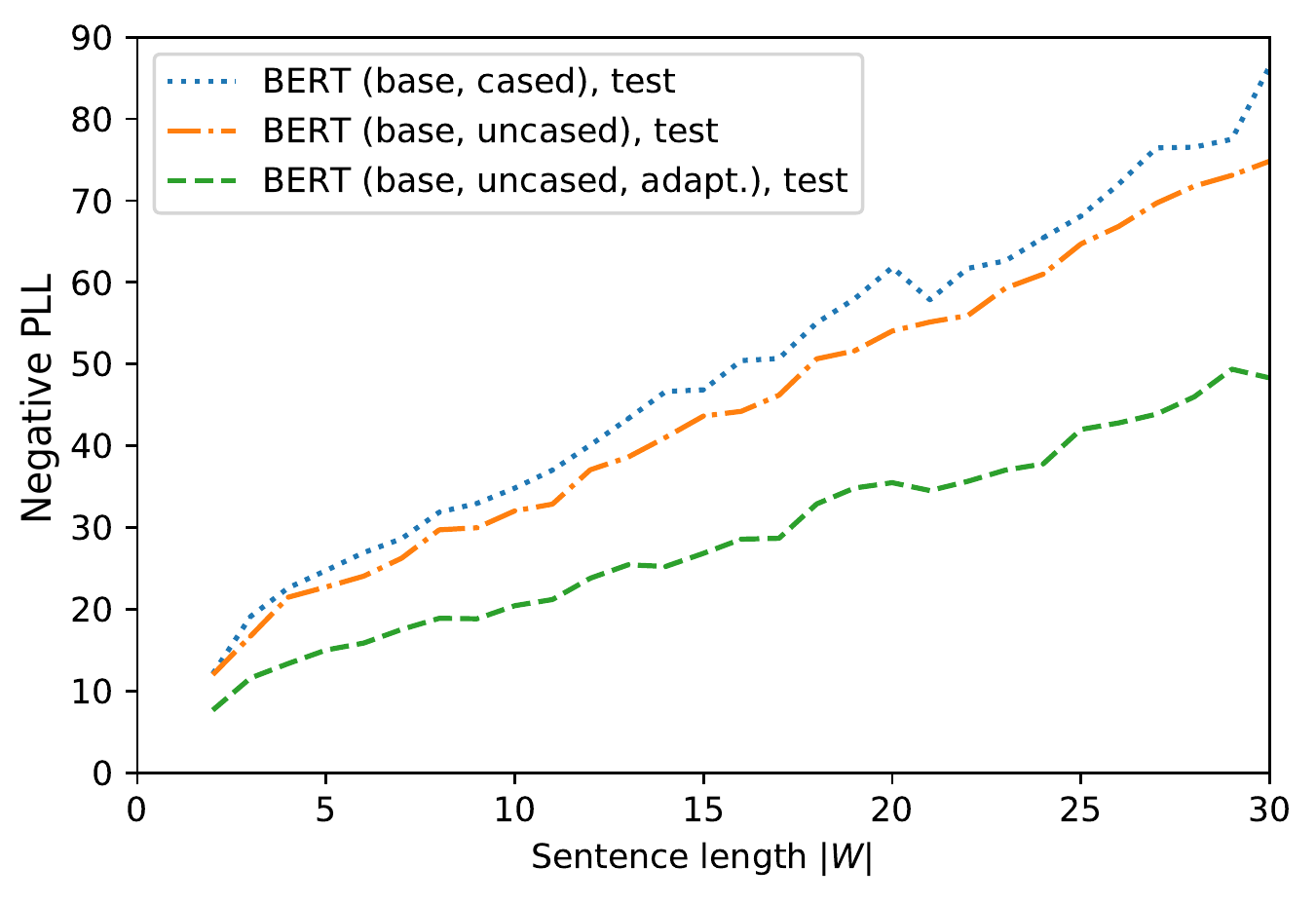}
\caption{Negative pseudo-log-likelihood scores versus sentence length (in tokens) from BERT, averaged over LibriSpeech's test utterances of each length.}
\label{fig:bert-pll}
\end{figure}

Note that in this section, we consider PPPLs normalized by \textit{number of words} (PPPL$_\text{w}$) to improve comparability between different subword vocabularies. We see a good correspondence between PPPL$_\text{w}$ improvements and post-rescoring WER in \Cref{table:intrinsic-ls}, and post-rescoring BLEU in \Cref{table:intrinsic-mt-en}.

\begin{table}[h!]
\begin{minipage}{1.0\linewidth}
	\centering
	\footnotesize
	\setlength{\tabcolsep}{4pt}
\begin{tabu}{@{}lcccc@{}}
\toprule
\multirow{3}{*}{\textbf{Model}} & \multicolumn{4}{c}{\textbf{test}} \\
& \multicolumn{2}{c}{clean} & \multicolumn{2}{c}{other} \\
& PPPL$_\text{w}$ & WER & PPPL$_\text{w}$ & WER \\
\midrule
BERT (base, cased) & 24.18 & 5.41 & 27.47 & 17.41 \\
RoBERTa (base, cased) & 21.85 & 5.25 & 24.54 & 17.18 \\
BERT (large, cased) & 17.49 & 5.25 & 19.59 & 16.97 \\
BERT (base, uncased) & 17.49 & 5.14 & 19.24 & 16.97 \\
RoBERTa (large, cased) & 14.78 & 5.05 & 16.23 & 16.79 \\
BERT (base, Libri. only) & 9.86 & 4.79 & 10.55 & 16.50 \\
BERT (base, unc., adapt.) & \textbf{6.63} & \textbf{4.58} & \textbf{6.56} & \textbf{15.96} \\
\bottomrule
\end{tabu}
\end{minipage}
\caption{Word-normalized PPPL vs.\ WER on LibriSpeech after rescoring, for models with different token vocabularies. WERs are from \Cref{table:wer-ls-pt} and \Cref{table:wer-ls-adapt}.}
\label{table:intrinsic-ls}
\end{table}
\begin{table}[h!]
\begin{minipage}{1.0\linewidth}
	\centering
	\footnotesize
\setlength{\tabcolsep}{3.5pt}
\begin{tabu}{@{}lcccccc@{}}
\toprule
\multirow{3}{*}{\textbf{Model}} & \multicolumn{6}{c}{\textbf{dev}} \\
& \multicolumn{2}{c}{ar\textrightarrow en} & \multicolumn{2}{c}{gl\textrightarrow en} & \multicolumn{2}{c}{sk\textrightarrow en}\\
& PPPL$_\text{w}$ & BLEU & PPPL$_\text{w}$ & BLEU & PPPL$_\text{w}$ & BLEU\\
\midrule
B-base & 13.08 & 35.71 & 11.86 & 20.25 & 13.20 & 29.74 \\
B-large & 10.17 & 35.79 & 9.48 & 20.21 & 10.43 & 29.79 \\
R-base & 9.77 & 35.86 & 9.36 & 20.21 & 9.75 & 29.79 \\
R-large & \textbf{6.26} & \textbf{36.02} & \textbf{6.08} & \textbf{20.44} & \textbf{6.29} & \textbf{30.05} \\

\bottomrule
\end{tabu}
\end{minipage}
\caption{Word-normalized PPPL vs.\ BLEU of cased BERT (B) and RoBERTa (R) on English gold sentences in the TED Talks corpus.}
\label{table:intrinsic-mt-en}
\end{table}

Thus, one could compute a new pretrained model's word-normalized PPPL on a small target-domain sample to quickly assess whether rescoring with it could improve on the previous model.

\section{Combining MLMs and GPT-2}
\label{ssec:combo}

We ask whether scores from a unidirectional LM are complementary with a masked LM for rescoring. When interpolating, we introduce $\gamma$ such that:
\begin{align*}
    \log g(\mW) = (1 - \gamma) \log P_{\text{LM}}(\mW) + \gamma\;\text{PLL}(\mW).
\end{align*}

Our results are in \Cref{table:wer-ls-interp}:

\begin{table}[h!]
\begin{minipage}{1.0\linewidth}
	\centering
	\footnotesize
\begin{tabu}{@{}lcccc@{}}
\toprule
\multirow{2}{*}{\textbf{Model}} & \multicolumn{2}{c}{\textbf{test}} & \multicolumn{2}{c}{\textbf{+ GPT-2}} \\
 & clean & other & clean & other \\
  \midrule 
baseline (100-best) & 7.26 & 20.37 & 5.30 & 17.26 \\
BERT (large, cased) & 5.25 & 16.97 & 5.03 & 16.80 \\
RoBERTa (large, cased) & 5.05 & 16.79 & 4.93 & 16.71 \\
BERT (base, unc., adapt.) & 4.58 & 15.96 & \textbf{4.50} & \textbf{15.92} \\
\bottomrule
\end{tabu}
\end{minipage}
\caption{WERs on LibriSpeech after hypothesis rescoring, with and without interpolating with GPT-2 (345M, cased).}
\label{table:wer-ls-interp}
\end{table}

As the MLM gets stronger, the improvement from adding scores from GPT-2 goes to zero, suggesting that their roles overlap at the limit. However, unlike recent work \citep{Shin2019} but like previous work \citep{Chen2017biLM}, we found that interpolating with a unidirectional LM remained optimal, though our models are trained on different datasets and may have an ensembling effect.

\end{document}